# Robust Principal Component Analysis Using Statistical Estimators

Peratham Wiriyathammabhum
Department of Computer Engineering
Chulalongkorn University
Bangkok, Thailand
u48pwy@cp.eng.chula.ac.th

Boonserm Kijsirikul
Department of Computer Engineering
Chulalongkorn University
Bangkok, Thailand
boonserm.k@chula.ac.th

**Abstract**

*Principal Component Analysis (PCA) finds a linear mapping and maximizes the variance of the data which makes PCA sensitive to outliers and may cause wrong eigendirection. In this paper, we propose techniques to solve this problem; we use the data-centering method and reestimate the covariance matrix using robust statistic techniques such as median, robust scaling which is a booster to data-centering and Huber M-estimator which measures the presentation of outliers and reweight them with small values. The results on several real world data sets show that our proposed method handles outliers and gains better results than the original PCA and provides the same accuracy with lower computation cost than the Kernel PCA using the polynomial kernel in classification tasks.*

## 1. Introduction

Nowadays, we have a very high potential for tasks in data collection which gives us data with many features to process such as gene data, image data, financial data, sensor data etc. Although we have a powerful computer which is equivalent to the old days super computer for processing, we still face the problem of the curse of the dimensionality due to the sparseness in the high-dimensional data we collect. The techniques for dimensional reduction are the solutions to avoid the curse of the dimensionality.

The techniques reduce the number of random variables in the data set and reduce a large to a smaller one. That means we can reduce computation cost and have a satisfied result because we still have the inner dimensions with essential parameters to process with minimized error.

Principal Component Analysis (PCA) [1] is a well known and widely used technique for dimensional reduction. The main idea of PCA is to reduce the dimension of the data set which has many relations among feature variables and retains data in maximum variance scheme by transforming the data set into principal components. Each feature in the principal component is not related and arranged by its importance so primary principal components can represent the variance of the data set.

However, PCA suffers from some limitations. To begin with, PCA uses a linear transformation so PCA does not work well on non-linear data sets. Moreover, principal components must be orthogonal to enable linear algebra to solve the problem. Finally, PCA assumes that large variance shows the behavior of the data and small variance shows noise. When there are outliers in the data set, PCA may mistake them as a behavior of the data instead of noise.

In this paper, we argue that we can improve PCA's estimators to another set of robust estimators to make PCA a robust PCA. It is shown that we can use median instead of mean and reweight the covariance matrix using M-Estimators such as Huber M-Estimator [2] which estimates on objective function defined by the size of each zero-meaned data.

## 2. Background

### 2.1 Dimensional Reduction

For a data set $X$ of dimension $D$, we denoted $x_1$ to $x_n$ for each data entry. For easy representation, we have an $n*D$ matrix as a data set. Assume that the data has the intrinsic dimension $d$ that can represent the data set ($d<D$). In other words, the data set $X$ is lying near a $d$ dimension manifold embedded in the $D$ dimension space [3]. Dimensional reduction techniques obtain the new data set $Z$ with the dimension d by finding a mapping from the original data $X$ to the underlying data $Z$.

## 2.2 Techniques for Dimensional Reduction

Nowadays, there are many dimensional reduction techniques developed in many ways that can be classified into two sets [3], i.e. linear techniques and nonlinear techniques, by identifying the linearity from the arrangement of components in the data set. Linear techniques, such as PCA or LDA, have gained popularity and are widely used. The main idea is to embedding the data to a lower dimension subspace by linear transformation. Several approaches for nonlinear techniques have also been proposed. Some of the approaches are graph-based techniques which extract the underlying embedding with graph construction, such as Isomap [4], LLE [5], Laplacian Eigenmaps [6], or kernel-based techniques which use kernel tricks to make linear techniques to be used in nonlinear scheme, such as Kernel PCA or Kernel LDA [7], and other techniques, such as Autoencoder [8] that uses neural networks to represent the lower dimension data by a middle hidden layer with d nodes. This paper studies robustness of linear-techniques for improvement of PCA.

## 3. Classical PCA

From a data set $X$ of dimension $D$ and the number of data $n$, we assume that $X$ is zero-meaned. We are trying to maximize the objective function.

$$\max_{U^T U=I} Tr(U^T X X^T U)$$

As covariance ($C$) equals $XX^T$ the objective function can be rewritten as:

$$\max_{U^T U=I} Tr(U^T C U).$$

This leads to the eigenvalue problem.

$$UC = \Delta U.$$

We use $d$ largest eigenvectors of $U$ ordered by magnitude of each corresponding eigenvalue $\Delta$ as the principal direction. Then, we transform the data set to principal components using principal directions as new bases which span a new subspace for the new data set $Z$.

This approach can be seen as using covariance to find the principal components. Another approach is to minimize the following objective function:

$$\min_{U,V} \|X - UV^T\|^2$$

where $V$ are the data projected on $U$. This is another approach that can be seen as approximating the low rank matrix from the original matrix and this approach also has the same solution with the first approach due to the linear algebra solver SVD, because we want the principal directions to be orthogonalized for effective result.

PCA finds new $d$ dimension bases for the new data space and makes a new data set by projecting original data to the new data space. This is the reason why PCA can preserve the structure of the data set. However, if the selected dimension is too low in cases of no redundant features, the result of PCA can be poor. As a result, we should know how to select the dimension $d$ effectively but this is not included in this paper.

## 4. Outlier, Robustness, Robust Estimator

### 4.1 Outliers

Outliers are data in the data set which can cause a surprise result when finding an underlying relation of the data. It is noted that outliers can be either a correct value or an incorrect one which can be made by any errors while inputting the data.

It is better to design outliers-tolerant algorithms than screening outliers from data for some reasons. To begin with, it is not fascinate to screening the data set every time. Moreover, the decision of which data is an outlier and screening it out is not as good as reweighting it with a low weight because rejecting all outliers can miss some interesting features in the data. Finally, it is difficult or sometimes impossible to spot the outliers from high-dimension data. Since the original data is in high dimension which are extremely hard to process, performing an outliers-detection process can suffer from the curse of dimensionality too.

### 4.2 Example of Robustness

Let explain the term robustness by an example. Given the data set of $n$ elements $x_1$ to $x_n$ with mean $\bar{x}$. Assume that the data values are numeric. If $x_n$ has a very large value compared with the rest, mean $\bar{x}$ can be induced by $x_n$ to make the mean value much more than most of the data and mislead the representation of the data set. However, if we use the median instead of the mean, the median is not affected by an outlier $x_n$ and can represent most data of the data set. The median can tolerate up to half of the data to be outlier which is clearly better than the mean. We say median is more robust than mean.

### 4.3 Robust Estimator

Estimations are means to use statistical methods, estimators, to approximate parameters. A good robust estimator has many properties. Firstly, it must be unbiased. A robust estimator should determine the estimate value near the real parameter. Secondly, it should be functional invariant which means that it can be used efficiently in any function. Finally, it should be asymptotically an unbiased mean that the biasness should converge to zero if the sample amount is large.

In this paper, we choose Huber M-Estimator [2] to modify the objective function. The Huber M-Estimator uses linear loss for outliers and square loss for other data to lower outliers' impact on error rate. However, users should determine the value $t$ which is the threshold value used to measure whether the data is an outlier. The Huber M-Estimator has a formula as follows.

$$\varepsilon_{Huber} = argmin_{\varepsilon \in \mathbb{R}^d} [\sum_{i=1}^{n} \rho(f(x_i) - y_i)]$$

$$\rho(y) = \begin{cases} \frac{y^2}{2} & : |y| \leq t \\ t|y| - \frac{t^2}{2} & : |y| > t \end{cases}$$

### 4.4 Scaling

Scaling [9] is a mean to boost the outlier detection-based method using location information of the data. There are two types of scaling which are auto scaling and robust scaling. Auto scaling uses standard deviation $s$ and mean $\bar{x}$ to modify the data which refers to normal distribution:

$$z_i = \frac{x_i - \bar{x}}{s}.$$

If the data has the distribution similar to normal distribution, we can estimate the number of data in the range of $z$ by both normal distribution scheme or by Chebychev's theorem. However, auto scaling also suffers from outliers because of using nonrobust estimator likes mean and standard deviation. This leads to another scaling method, robust scaling. Robust scaling uses median instead of mean and standard absolute deviation instead of standard deviation. Robust scaling has many approaches. In this paper, we used two types of robust scaling which are for symmetric distributed data ($S_{mad}$) and asymmetric distributed data ($S_n$):

$$S_{mad} = 1.4826 med_i(|x_i - med_j(x_j)|)$$
$$S_n = 1.1926 med_i(med_j(|x_i - x_j|)).$$

## 5. Proposed Algorithm

Our proposed method, DC-HPCA, uses the data-centering method to detect outliers. Data-centering uses Euclidean distance from the center of the data, assumed to be the origin point after preprocessing, for the criteria of measuring outliers. The preprocessing is robust scaling. With robust scaling, we can choose the percentile of the data assumed to be data or outliers with an input parameter. Then, we use the Huber M-Estimator to weight the covariance matrix of the scaled data. Finally, we solve the eigenvalue problem and obtain the eigen vectors sorting in ascending order because we weight them in terms of errors. The outline algorithm is shown in Fig.1.

| |
|---|
| Input: Array of numeric multivariate data $X$ size $m*n$, lower dimension of positive integer $d$, the percentile parameter of integer value $c$ ($0 \leq c \leq 100$) |
| Output: Array of dimension reduced data $Z$ size $m*d$, array of eigenvectors $U$ size $n*d$, array of eigenvalues $\triangle$ size $d*1$. |
| STEP 1: Compute the median of each parameter of $X$. <br> STEP 2: Find the standard absolute deviation $S_{mad}$ or $S_n$ $$S_{mad} = 1.4826 med_i(|x_i - med_j(x_j)|)$$ $$S_n = 1.1926 med_i(med_j(|x_i - x_j|)).$$ STEP 3: Do the robust scaling $$z_i = \frac{x_i - \bar{x}}{s}$$ where $\bar{x}$ is a median and $s$ is $S_{mad}$ or $S_n$. <br> STEP 4: Find value $t$ for Huber M-Estimator from the size of the data sorted in descending order and choose by using percentile parameter $c$. <br> STEP 5: Compute the covariance matrix $C$ weighting by Huber M-Estimator $$\rho(y) = \begin{cases} \frac{y^2}{2} & : |y| \leq t \\ t|y| - \frac{t^2}{2} & : |y| > t \end{cases}.$$ STEP 6: Solve the eigenvalue problem from matrix $C$. <br> STEP 7: Perform the dimensional reduction by choosing $d$ eigenvectors from $U$ in ascending order of eigenvalues $\triangle$. |

*Figure 1. The proposed algorithm*

# 6. Experimental Result

## 6.1 Synthetic Dataset

We tested the effect of outliers on the original PCA and our proposed method. In Fig.2, we synthesized 33 data points forming on the vector *<0.707,0.707>* and 3 of 33 points are outliers. We found that PCA is affected by outliers significantly noticed by the wrong eigen directions that did not pass on most data. For robust methods, HPCA (PCA with Huber M-Estimator) and R1PCA [10] obtained the desired eigen directions along most data. Our two methods performed like them but were distinct on the second direction.

## 6.2 UCI Repository Datasets

We perform 10-fold cross validation with our proposed method on 5 UCI Datasets [11] comparing the classification performance of 1-Nearest Neighbors Classifier (1-NN) with PCA, KPCA with both Gaussian RBF Kernel and Polynomial Kernel, LPP [12] and two of robust PCA techniques, R1PCA [10] and Simple PCA [13]. A summary of 5 datasets is in Table 1 and the classification results are shown in Table 2.

The results indicate that our proposed method provides classification accuracy comparable to the Kernel PCA. However, our method is an extension of PCA in term of robustness so it is a linear technique with low computation cost. When compared with other methods, our method has an advantage in term of classification accuracy using a 1-NN classifier and is easier to implement; however, R1-PCA and SPCA require less time and memory than our method.

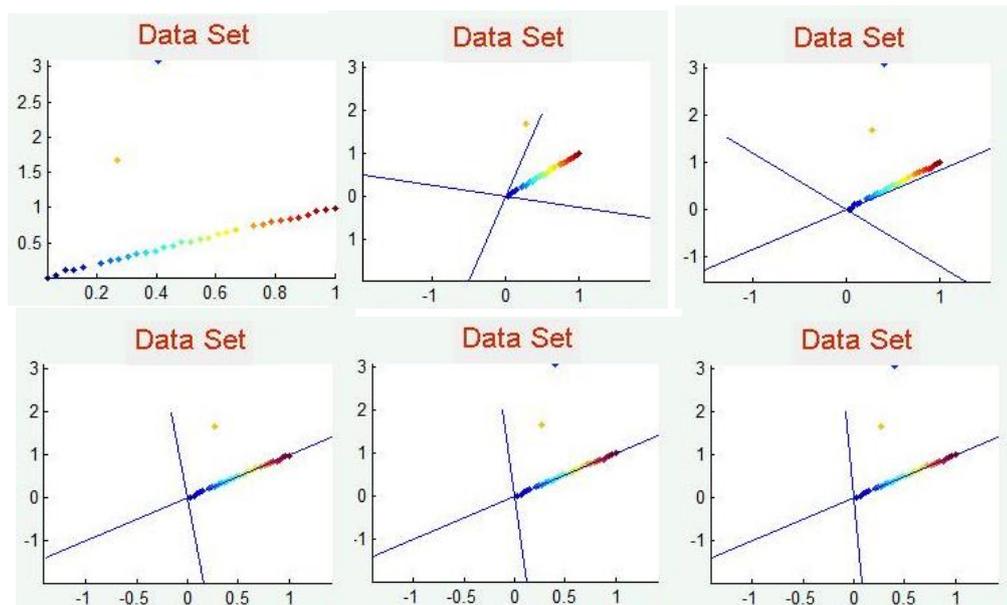

Figure 2. First column: Straight line-like dataset with 3 outliers, Eigendirections of PCA, R1-PCA
Second column: Eigendirections of HPCA, DC-HPCA($S_{mad}$), DC-HPCA($S_n$) (from left to right).

TABLE I. UCI DATASETS USED IN THE EXPERIMENT

| Dataset | Dimension | Number of classes | Number of Instances |
|---|---|---|---|
| Ionosphere | 34 | 2 | 351 |
| Tic-Tac-Toe Endgame | 9 | 2 | 958 |
| Wine | 13 | 3 | 178 |
| Parkinsons | 23 | 2 | 197 |
| Glass Identification | 10 | 7 | 214 |

TABLE II. CLASSIFICATION ACCURACY OF UCI DATASETS USING 1-NN

| Dataset | D | PCA | KPCA (Gauss) | KPCA (Poly) | LPP | R1PCA | SPCA | DC-HPCA $S_{mad}$ | DC-HPCA $S_n$ |
|---|---|---|---|---|---|---|---|---|---|
| Ionosphere | 2 | 71.51 | 80.06 | **80.63** | - | 72.36 | 77.49 | 76.64 | 72.93 |
|  | 3 | 85.47 | **91.43** | 84.05 | - | 76.92 | 82.05 | 84.05 | 78.92 |
| Tic-Tac-Toe Endgame | 2 | 58.25 | 58.14 | 96.66 | 65.34 | 97.08 | 97.60 | 98.54 | **98.75** |
|  | 3 | 59.29 | 61.59 | 96.66 | 63.26 | 95.72 | 97.60 | **98.85** | 98.85 |
| Wine | 2 | 72.47 | 35.39 | **74.16** | 42.13 | 71.91 | 70.22 | 61.80 | 68.54 |
|  | 3 | 73.03 | 38.20 | 73.60 | 39.89 | 73.03 | 72.47 | **76.40** | 73.03 |
| Parkinsons | 2 | 84.62 | 65.13 | **86.67** | 64.10 | 80.00 | 78.46 | 82.56 | 84.10 |
|  | 3 | **84.62** | 66.15 | **84.62** | 67.69 | 78.97 | 82.56 | **84.62** | **84.62** |
| Glass Identification | 2 | 97.66 | 40.65 | 97.20 | 38.79 | 98.13 | 95.33 | 91.12 | **98.60** |
|  | 3 | 97.66 | 41.59 | **98.60** | 42.06 | **98.60** | 96.73 | 94.39 | **98.60** |

## 7. Summary

In this paper, we have proposed an application of data-centering with robust scaling and Huber M-Estimator to principal component analysis for outliers handling. The proposed method tries to find a circle that can control the number of data in it and weights the data out of this circle to be outliers. We use robust scaling to scale the data in normal distribution scheme and make the circle to be more precise on outliers detection. The experimental results show that our proposed method (DC-HPCA $S_n$) provides better performance than the original PCA because most of the real world data are distributed in asymmetric scheme.